\documentclass[10pt,twocolumn,letterpaper]{article}

\usepackage{iccv}
\usepackage{times}
\usepackage{epsfig}
\usepackage{graphicx}
\usepackage{amsmath}
\usepackage{amssymb}
\usepackage{bm}
\usepackage{float}
\usepackage{caption}
\usepackage{subcaption}
\usepackage{multirow}
\usepackage{color, colortbl}

\usepackage[pagebackref=true,breaklinks=true,letterpaper=true,colorlinks,bookmarks=false]{hyperref}

\iccvfinalcopy 

\def\iccvPaperID{106} 

\definecolor{Gray}{gray}{0.9}
\definecolor{LightCyan}{rgb}{1, 1, 0.8}

\ificcvfinal\fi
\begin{document}

\title{Scene Labeling using Gated Recurrent Units with Explicit Long Range Conditioning}

\author{Qiangui Huang\\
University of Southern California\\
Los Angeles, CA\\
{\tt\small qianguih@usc.edu}
\and
Weiyue Wang\\
University of Southern California\\
Los Angeles, CA\\
{\tt\small weiyuewa@usc.edu}
\and
Kevin Zhou\\
Siemens Healthineers\\
Princeton, NJ\\
{\tt\small s.kevin.zhou@gmail.com}
\and
Suya You\\
University of Southern California\\
Los Angeles, CA\\
{\tt\small suya@usc.edu}
\and
Ulrich Neumann\\
University of Southern California\\
Los Angeles, CA\\
{\tt\small uneumann@usc.edu}
}

\maketitle

\begin{abstract}
   Recurrent neural network (RNN), as a powerful contextual dependency modeling framework, has been widely applied to scene labeling problems. However, this work shows that directly applying traditional RNN architectures, which unfolds a 2D lattice grid into a sequence, is not sufficient to model structure dependencies in images due to the ``impact vanishing'' problem. First, we give an empirical analysis about the ``impact vanishing'' problem. Then, a new RNN unit named Recurrent Neural Network with explicit long range conditioning (RNN-ELC) is designed to alleviate this problem. A novel neural network architecture is built for scene labeling tasks where one of the variants of the new RNN unit, Gated Recurrent Unit with Explicit Long-range Conditioning (GRU-ELC), is used to model multi scale contextual dependencies in images. We validate the use of GRU-ELC units with state-of-the-art performance on three standard scene labeling datasets. Comprehensive experiments demonstrate that the new GRU-ELC unit benefits scene labeling problem a lot as it can encode longer contextual dependencies in images more effectively than traditional RNN units.

   
\end{abstract}

\section{Introduction}

\begin{figure}
\begin{center}
   \includegraphics[width=1\linewidth]{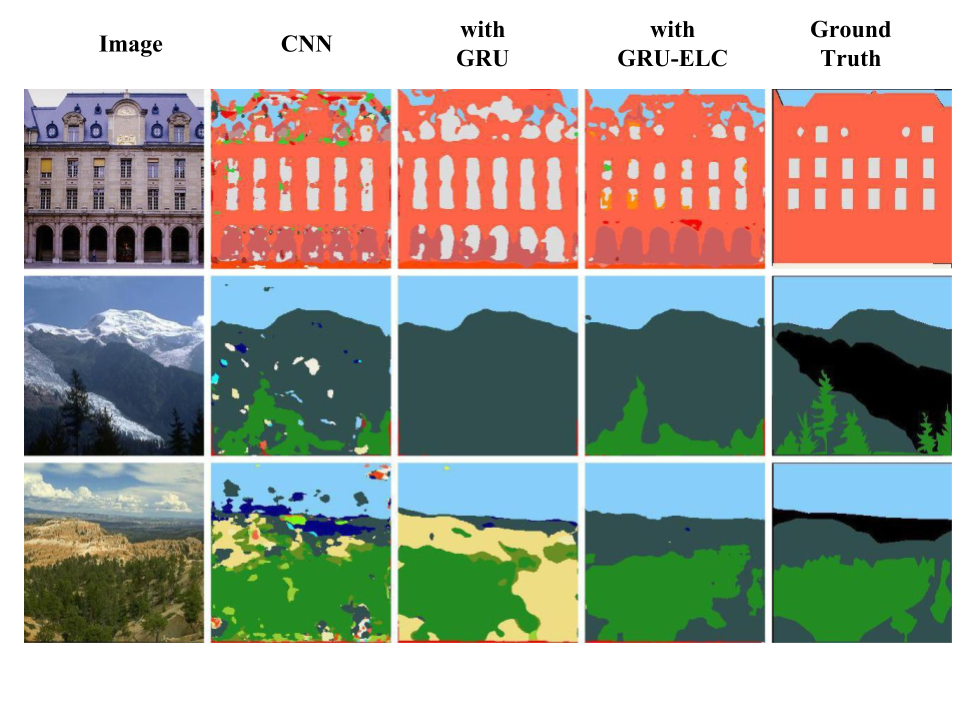}
\end{center}
   \caption{CNNs have challenges in dealing with local textures in images as shown in the second row. With the help of Gated Recurrent Units, the model can make globally better prediction. However, GRUs still struggle in modeling fine structures in images due to the ``impact vanishing'' problem. Our GRU-ELC units can effectively model multi scale contextual dependencies in images and thus successfully preserve local details in predictions, such as the windows and doors (even not annotated in ground truth) in first row, and the trees and mountains in the second and third row.}
\label{fig:siftflow_short}
\end{figure}

Scene labeling is a fundamental task in computer vision. Its goal is to assign one of many pre-defined category labels to each pixel in an image. It is usually formulated as a pixel-wise multi-class classification problem. Modern scene labeling methods rely heavily on convolutional neural networks (CNNs) \cite{2,3,8,9}. CNNs are capable of learning scale-invariant discriminative features for images. These features have proven more powerful than traditional hand-crafted features on many computer vision tasks \cite{43,44,3}. Specially designed CNN architectures \cite{2,3,5,6,8} have shown superior performance on scene labeling by using end-to-end training. 

However, CNNs have challenges in dealing with local textures and fine structures in images and tend to over-segment or under-segment objects in images. To accurately segment small components and detect object boundaries, long range contextual dependencies in images are usually desired when designing scene labeling algorithm. 
Many works have exploited probabilistic graphical models such as conditional random fields (CRFs) \cite{22} to capture structural dependencies in images \cite{8,9}. However, CRFs usually require carefully designed potentials and their exact inference is usually intractable. In contrast, recurrent neural networks (RNN), as another category of powerful contextual dependency modeling methods, are free from these disadvantages and can learn contextual dependencies in a data-driven manner.

RNNs have first proven effective in modeling data dependencies in domains such as natural language processing and speech recognition \cite{11,12}. Recently, there are some attempts at applying RNNs to images \cite{10,14,16,17}. Because the spatial relationship among pixels in 2D image data is fundamentally different from the temporal relationship in the 1D data in NLP or speech recognition, variants of RNN architectures \cite{23, 16, 14, 24, 25, 26, 28} have been proposed to handle 2D image data, which typically involve unfolding a 2D lattice grid into a 1D sequence. The unfolded 1D sequence is usually much longer than the data sequence in NLP or speech recognition. Take a feature map of size $64\times64$ for example. Its unfolded 1D sequence is of length $64\times64=4096$. One major flaw of applying existing RNN units to sequences of such a long length is that the ``impact vanishing'' problem will raise and break the spatial dependencies in images.

It is well-known that long term dependency is hard to learn in RNN units due to the gradient exploding or vanishing problems and RNN units such as Long Short Term Memory (LSTM) or Gated Recurrent Unit (GRU) can effectively avoid these problems. However, in this work, we empirically show that even in LSTMs or GRUs, the dependency in a extremely long range is still hard to capture due to the ``impact vanishing'' problem.

First, this paper studies the ``impact vanishing" problem when traditional RNNs are applied to image data. Then, we generalize traditional RNN units to RNN units with Explicit Long-range Conditioning (RNN-ELC) to overcome this problem. Specifically, two variants, GRU-ELC and LSTM-ELC, are designed and discussed.

Compared with existing works \cite{23, 16, 14, 24, 25, 26, 28}, RNN-ELCs can effectively capture long range contextual dependency in images with the help of their explicit long range conditioning architecture. In the RNN-ELC units, the present variable is explicitly conditioned on multiple contextually related but sequentially distant variables. Intuitively, it is adding skip connection between hidden states. Adding skip connections in RNNs to help learn long term dependency first appears in \cite{33}. Our method generalizes the idea to model more complex 2D spatial dependencies. The RNN-ELC unit is applicable to both raw pixels and image features. It can be naturally integrated in CNNs, thereby enabling joint end-to-end training. 

In order to take the benefits of the new RNN units for scene labeling tasks, we build a novel scene labeling system using the GRU-ELC units to model long range multi-scale contextual dependencies in image features.

In summary, our main contributions include:
1) An empirical study of the ``impact vanishing" problem, which commonly exists in traditional RNN units when they are applied to images; 
2) A new RNN unit with Explicit Long-range Conditioning to alleviate the ``impact vanishing" problem;
3) A novel scene labeling algorithm based on GRU-ELCs. There are a few works utilizing GRUs for scene labeling. However, we show that our GRU-ELC units can actually achieve state-of-the-art performances in scene labeling tasks; and
4) Improved performances on several standard scene labeling datasets.

\section{Related Work}

Scene labeling is one of the most challenging problems in computer vision. Recently, convolutional neural network based methods achieved great success in this task. Farabet \textit{et al.} \cite{1} made one of the earliest attempts at applying hierarchical features produced by CNNs to scene labeling. Eigen \textit{et al.} \cite{2} designed a multi-scale convolutional architecture to jointly segment images, predict depth, and estimate normals for an image. Long \textit{et al.} \cite{3} applied fully convolutional network (FCN) to this task. Noh \textit{et al.} \cite{4} also used deconvolution layers for image segmentation. They adopted an encoder-decoder architecture, where encoder part consists of convolution and pooling operations and decoder part consists of deconvolution and unpooling operations. Badrinarayanan \textit{et al.} \cite{5} designed a similar architecture named SegNet. In \cite{6}, Yu and Koltun developed a dilated convolutional module to preserve multi-scale contextual information for image segmentation.

Although CNN based methods introduced powerful scale-invariant features for scene labeling, they performed poorly in preserving local textures and fine structures in predictions. These problems were addressed by combining CNNs with probabilistic graphical models such as conditional random fields (CRFs). Chen \textit{et al.} \cite{7} suggested to put a fully connected CRF \cite{22} on top of FCN to capture structural dependencies in images. Zheng \textit{et al.} \cite{8} showed that CNN and CRF can be jointly trained by passing the inference errors of CRFs back to CNN. Liu \textit{et al.} \cite{9} improved \cite{8} by introducing a more complex pairwise term for CRF. CRF based methods usually require carefully designed pair-wise potentials and unfortunately their exact inference is usually intractable.

RNNs, as another powerful tool for modeling contextual dependences in data, have achieved tremendous success in many areas such as speech recognition \cite{11} and natural language processing \cite{12}. There is also a rich literature of using RNNs for image related tasks \cite{17,18, 19, 20, 21}. 

Liang \textit{et al.} \cite{27} designed a graph LSTM to handle image data. However, their method is built on superpixels, which is computationally expensive and is not directly applicable to image features. Byeon \textit{et al.} \cite{10} developed a scene labeling method based on a 2D LSTM network, which first divided an input image into non-overlapping patches and then sequentially fed them into LSTM networks in four different orders. \cite{13} built a RNN segmentation algorithm based on recently proposed ReNet \cite{14}. Their idea was to to alternatively sweep an image in different directions and then sequentially input each row (or column) into a RNN. Shuai \textit{et al.} \cite{15,16} designed a quaddirectional 2D RNN architecture for scene labeling, where each pixel was connected to its 8 nearest neighbors.

Long range contextual dependencies are usually desired for scene labeling tasks. In previous works \cite{23, 16, 14, 24, 25, 26, 28}, the present variable only explicitly conditions on variables within a short range such as its 8 nearest neighbors. However, as shown later, short range conditioning does not capture long range structural dependencies in images. Our labeling algorithm is a generalized version of \cite{13,15,16}. But our model is different with them in that (i) our model is built upon a new RNN unit, GRU-ELC unit, which is free of ``impact vanishing'' problem; (ii). our models can effectively utilize multi scale contextual dependencies to provide better scene labeling performances.

\section{Recurrent Neural Networks with Explicit Long-range Conditioning}
\subsection{Recurrent Neural Network}
A vanilla RNN unit has two types of dense connections, namely, input-to-hidden and hidden-to-hidden connections. At each time step $t$, the output $y^{t}$ conditions on the input at current time step $x^{t}$ and the hidden state at previous time step $h^{t-1}$. Mathematically, given a sequence of input data $X=\{x^{t}, t=1,...,T\}$, the vanilla RNN unit models the probability of current time step output, $P( y^t | x^t, h^{t-1}, \theta)$, by the following equations:

\begin{gather}
  h^{t} = \sigma_{h}({x^{t}W_x + h^{t-1}W_h+b_h} ),
  \\
  y^{t} = \sigma_{y}({h^{t}W_y + b_y}),
\end{gather}

\noindent
where $\theta = \{W_{x,h,y},b_{h,y}\}$ is the parameter and $\sigma_h$ and $\sigma_y$ are the nonlinearity functions of hidden and output layers, respectively. $y^t$ explicitly conditions only on $x^t$ and $h^{t-1}$, but $h^{t-1}$ explicitly conditions on previous input $x^{t-1}$ and hidden state $h^{t-2}$; thus, $y^t$ actually implicitly conditions on all previous inputs and hidden states. Therefore, previous variables can influence their following variables by passing information through the hidden states.

However, vanilla RNNs have the notorious gradient vanishing or exploding problem when they are applied to learn long-term dependencies \cite{12, 29}. In practical applications, two types of $gated$ RNNs are developed to avoid this problem, Long Short-Term Memory (LSTM) network \cite{30} and Gated Recurrent Unit (GRU) network \cite{31}. 

LSTM uses the following equations to update its hidden states. Let $i^t$, $f^t$, $c^t$, and $o^t$ denote the output of the input gate, the forget gate, the cell gate, and the output gate, respectively. $\odot$ denotes element-wise multiplication. $W_{xi,xf,xc,xo,hi,hf,hc,ho}$, $w_{ci,cf,co}$, and $b_{i,f,c,o}$ are parameters. $\sigma_{i,f,c,o}$ are nonlinearity functions.

\begin{gather}
   i^t = \sigma_i (  x^t W_{xi} + h^{t-1} W_{hi} + w_{ci} \odot c_{t-1} + b_i       )    \\
   f^t = \sigma_f (  x^t W_{xf} + h^{t-1} W_{hf} + w_{cf} \odot c_{t-1} + b_f       )    \\
   c^t =  f_t \odot c_{t-1} + i_t \odot \sigma_c ( x^t W_{xc} + h_{t-1}W_{hc} + b_c  )  \\
   o^t = \sigma_o (  x^t W_{xo} + h_{t-1} W_{ho} + w_{co} \odot c_t + b_o  ) \\
   h^t = o^t \odot \sigma_h (c^t)
\end{gather}

GRUs compute the output by the following equations with $W_{xr,hr,xu,hu,xc,hc}$ and $b_{r,u,c}$ being parameters.

\begin{gather}
   r^t = \sigma_r (  x^t W_{xr} + h^{t-1} W_{hr} + b_r       )    \\
   u^t = \sigma_u (  x^t W_{xu} + h^{t-1} W_{hu} + b_u       )    \\   
   c^t = \sigma_c (  x^t W_{xc} +  r^t \odot (  h^{t-1} W_{hc} ) + b_c       )    \\
   h^t = (1-u^t) \odot h^{t-1} + u^t \odot c^t
\end{gather}

LSTMs and GRUs have been widely used in modeling long term dependencies as they can effectively prevent gradients from vanishing or exploding \cite{32}. However, they are still limited for image related applications because the ``impact vanishing'' is hard to avoid when dealing with a very long sequence.. 

\subsection{Impact Vanishing Problem}

\begin{figure}
  \centering
  \begin{tabular}[b]{c}
    \includegraphics[width=0.4\linewidth]{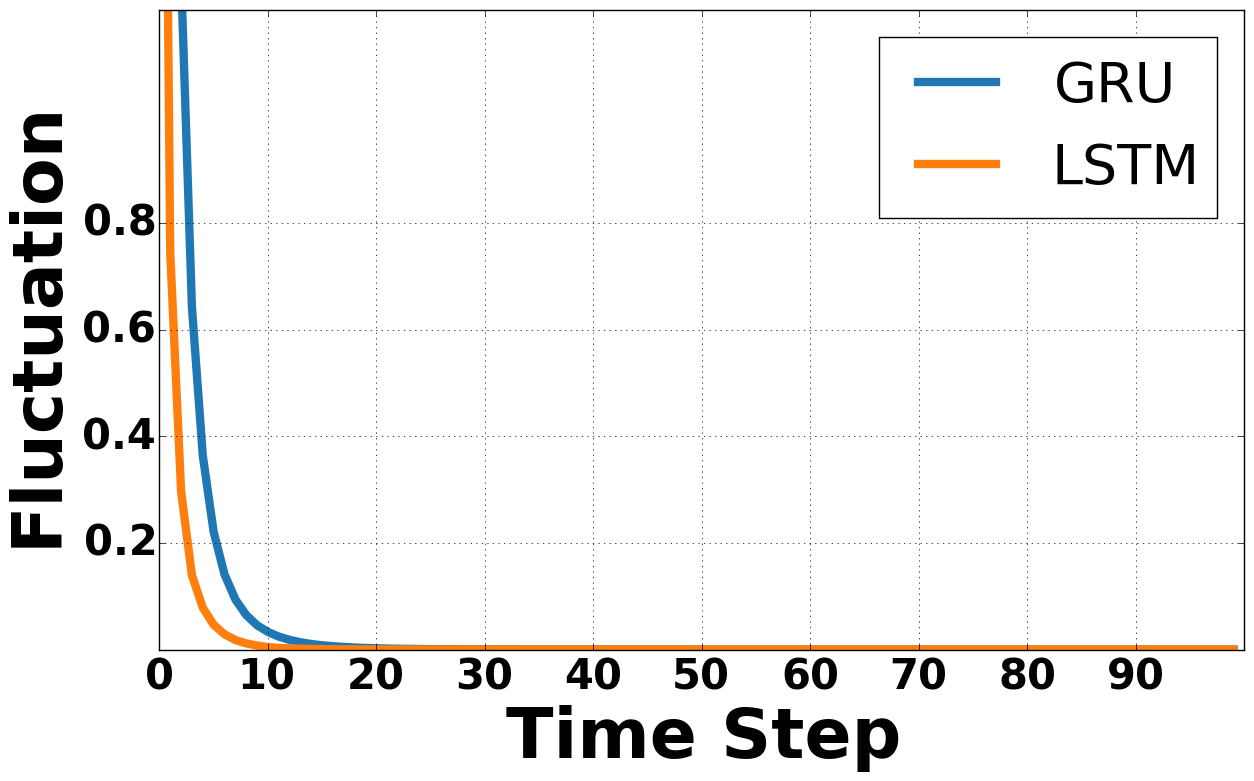} \\
    \small (a)
  \end{tabular} \qquad
  \begin{tabular}[b]{c}
    \includegraphics[width=0.4\linewidth]{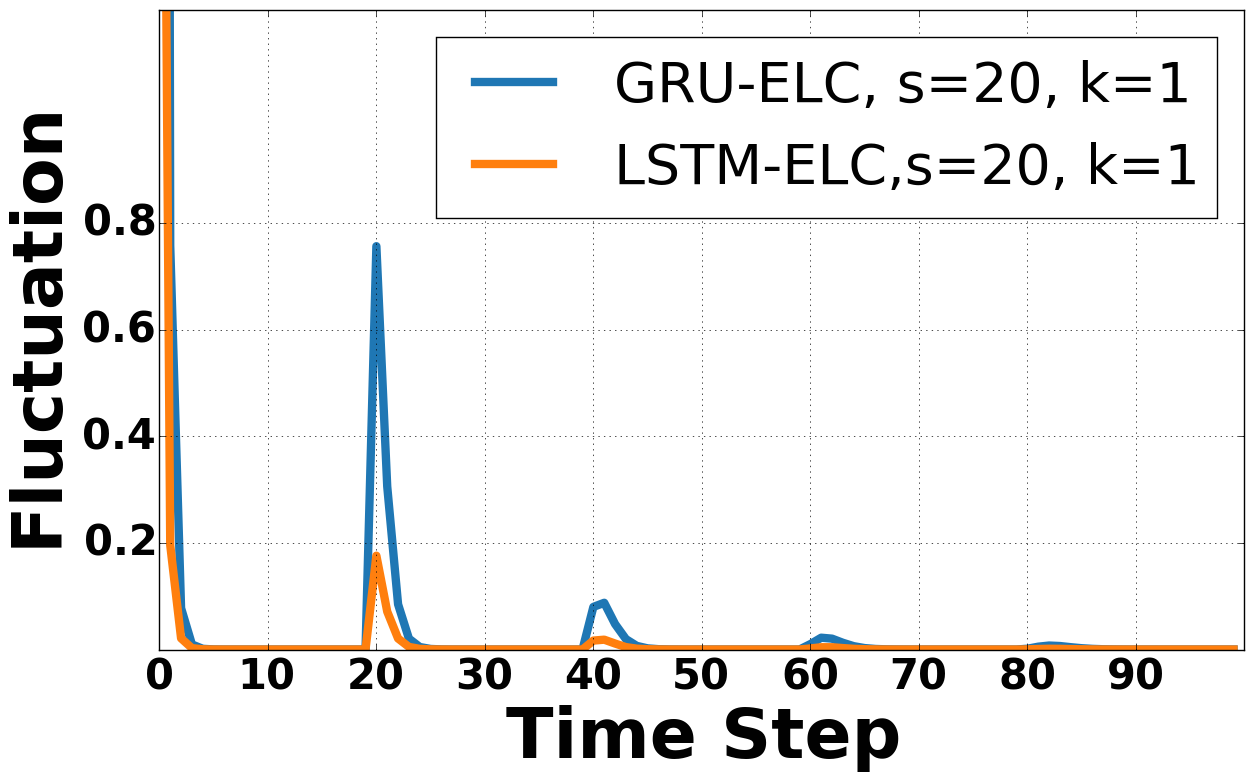} \\
    \small (b)
  \end{tabular} \qquad
      \begin{tabular}[b]{c}
    \includegraphics[width=0.4\linewidth]{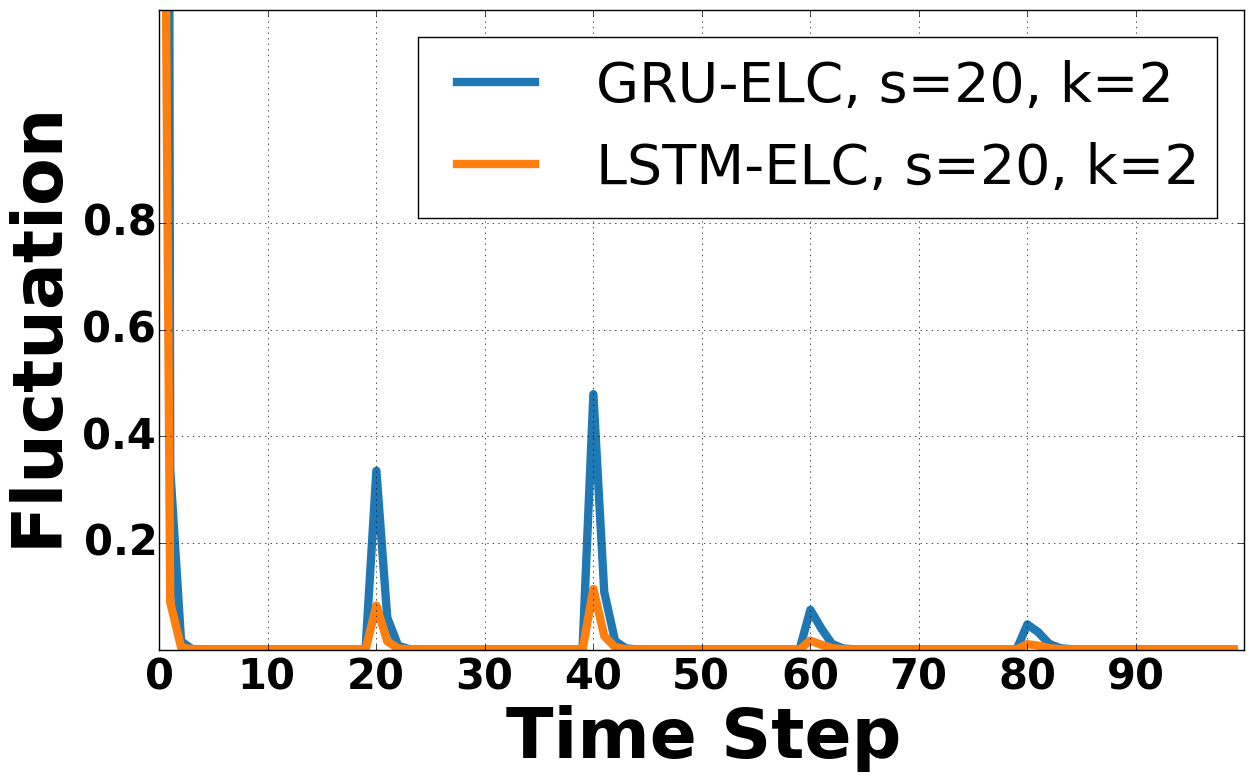} \\
    \small (c)
  \end{tabular} \qquad
    \begin{tabular}[b]{c}
    \includegraphics[width=0.4\linewidth]{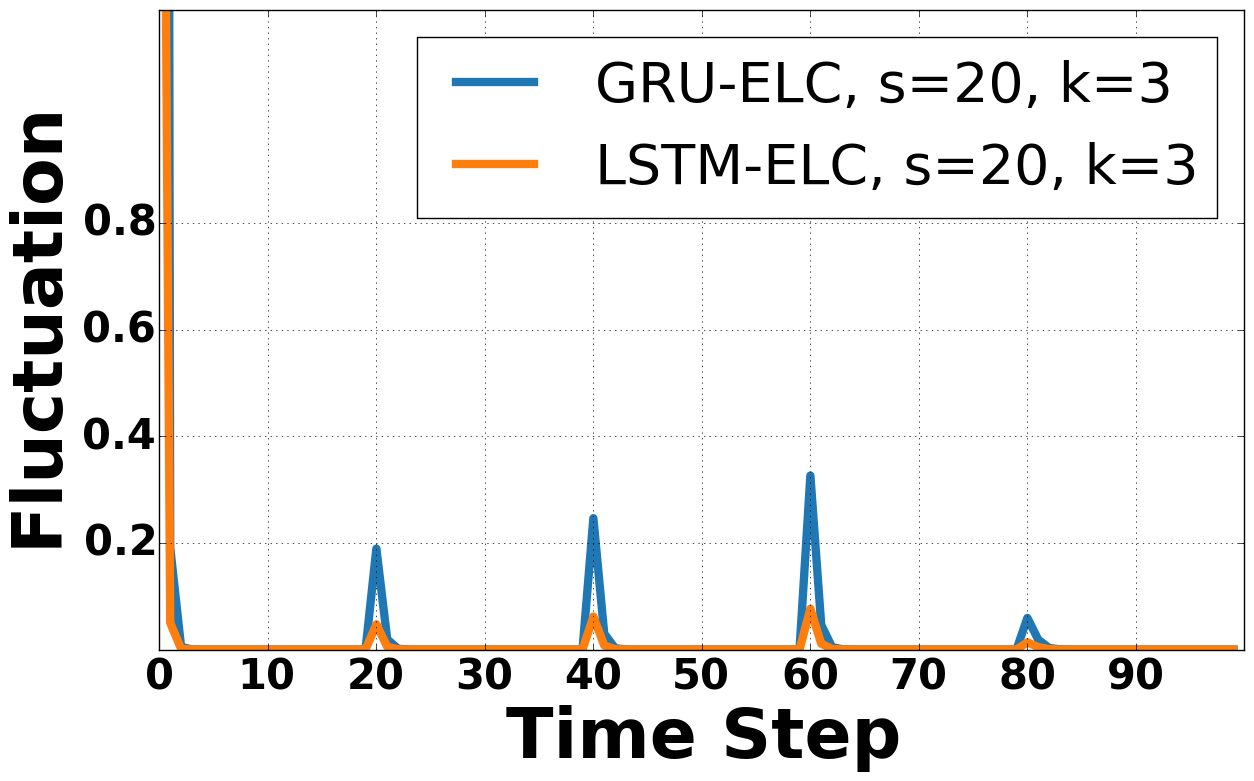} \\
    \small (d)
  \end{tabular}

  \caption{Illustration of ``impact vanishing" problem in LSTMs and GRUs. Fluctuation denotes the difference caused by changing the first data $x^1$ in input sequence.}
  \label{fig:iv}
\end{figure}

Here we design a concise and straightforward toy example to empirically demonstrate the existence of ``impact vanishing'' problem in LSTM/GRU units.

Assume $X=\{x^{t}, t=1,...,T\}$ is an input sequence. $x^t  \in R^{M \times N}$ is the input data at time step $t$. In this toy example,  $x^t$ is generated from a continuous uniform distribution $ U (0,1)$ and all weight parameters $W$ in LSTMs and GRUs are initialized with a Guassian distribution $\mathcal{N}(0,0.1)$ and bias parameters are set to zero. 

Firstly, the entire sequence $X$ is fed to LSTMs/GRUs which output $Y=\{y^{t}, t=1,...,T\}$. Then, the data at the first time step $x^1$ is replaced with $\hat{x}^1$ which is generated from same distribution $ U (0,1)$ and get a new input data sequence $\hat{X} = \{ \hat{x}^1, x^2,...,x^T   \}$. Feed $\hat{X}$ to the same RNNs and get the new output $\hat{Y}=\{ \hat{y}^1, \hat{y}^2,...,\hat{y}^T \}$. If the information of the first input can successfully pass through hidden states and make impact on following variables, we should expect $\hat{y}^t$ to be different with $y^t$ when $x^1$ changes. To measure how different $\hat{y}^t$ becomes, we calculate the following fluctuation metric:

\begin{gather}
  F^t = \frac{1}{MN} \sum_{i=1,j=1}^{M,N} ( y^t(i,j) - \hat{y}^{t}(i,j) ) ^2
\end{gather}

We repeat this process by 20 times, collect all $F^t$, and report the mean of all 20 $F^t$ in the top left plot in Fig.\ref{fig:iv}. It shows that  $F^t$ drops dramatically in the first 10 time steps. When $t>20$, $F^t$ decreases to zero, which means that $y^t$ stays unchanged when $t>20$ regardless of the initial input $x^1$. \textbf{Although $h^t$ implicitly depends on $h^1$ and $x^1$, the dependency between the $1^{th}$ and $t^{th}$ variables is actually broken when $t>20$}. In another word, $P(h^t|h^{t-1},..., h^1, h^0) = P(h^t|h^{t-1},..., h^1)$ when $t>20$. This phenomenon is referred as ``impact vanishing" problem in this paper. 

The mechanics behind ``impact vanishing'' problem is similar to the gradient vanishing/exploding problem \cite{12}. The $t^{th}$ variable makes impact on following variables by storing its information in $h^t$ and passing it to following hidden states. During the flow, $h^t$ will be multiplied by fixed weight matrixes many times. If the spectral radius of the weight matrix is smaller than 1, the multiplication results will vanish to zero. And if the spectral radius is larger than 1, multiplication results will explode to infinity and thus saturates the sigmoid and tanh functions. In both cases, information stored in $h^t$ has decreasing impact on $h^{t+k}$ when $k$ becomes larger.

 \begin{figure}[t]
\centering
\begin{subfigure}{.47\textwidth}
  \centering
  \includegraphics[width=.2\linewidth]{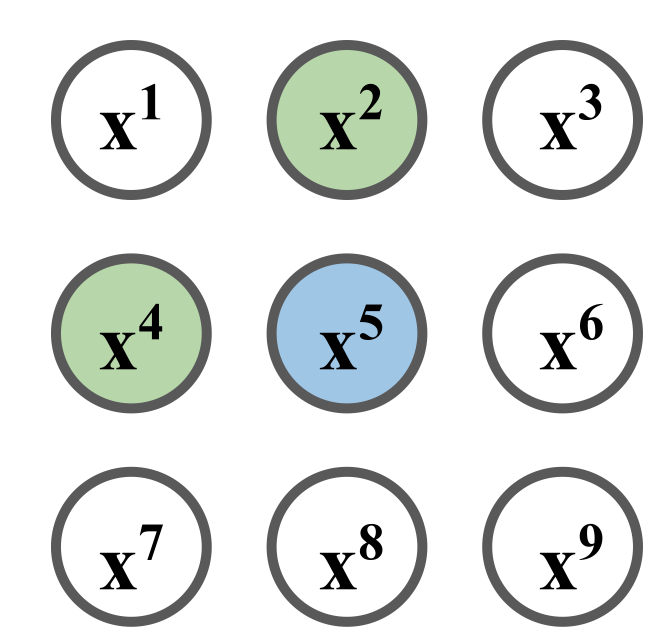}
  \caption{Original 2D image grid}
    \label{fig:image_grid}
\end{subfigure}%


\begin{subfigure}{.47\textwidth}
  \centering
  \includegraphics[width=0.8\linewidth]{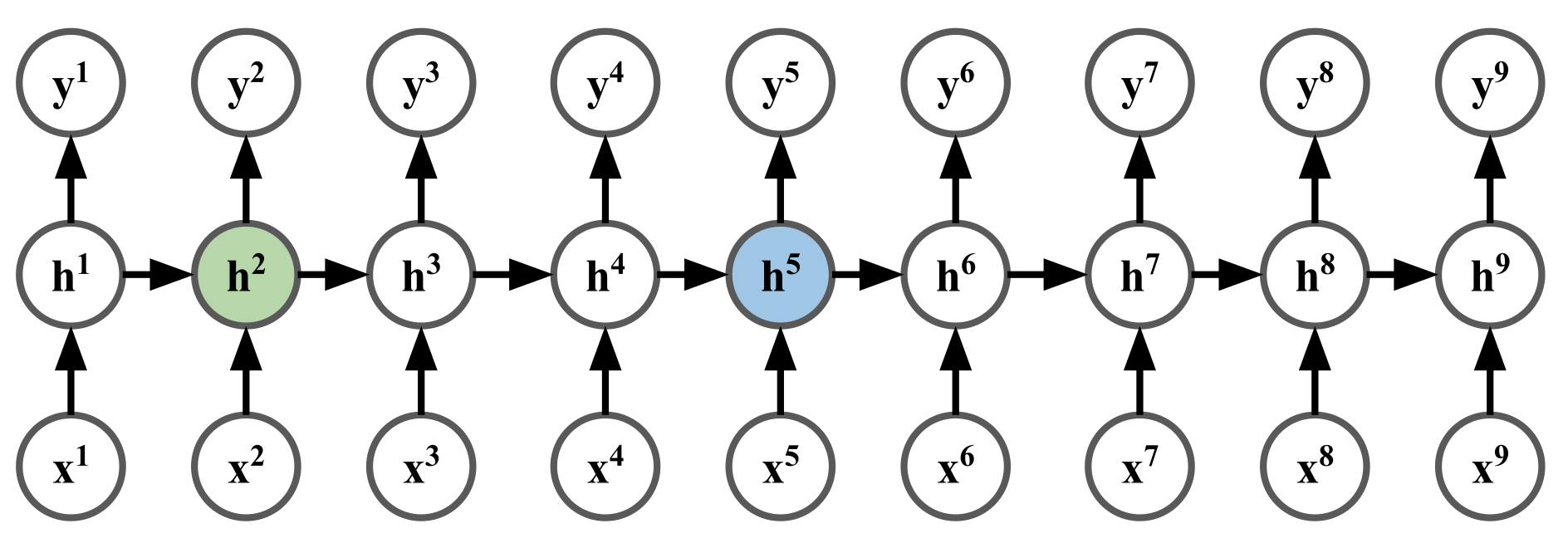}
  \caption{Unfolded image grid modeled by traditional RNN unit}
  \label{fig:unfold}
\end{subfigure}

\begin{subfigure}{.47\textwidth}
  \centering
  \includegraphics[width=0.8\linewidth]{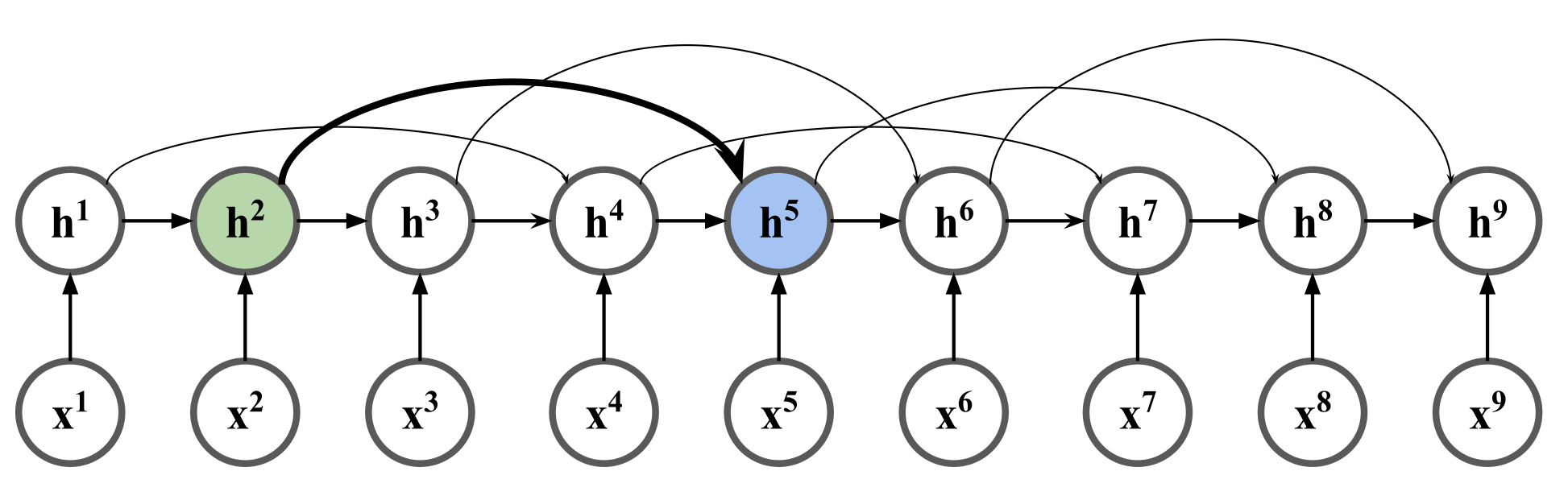}
  \caption{Unfolded image grid modeled by RNN-ELC units}
      \label{fig:unfold_skip}
\end{subfigure}

   \caption{Graphical illustration of unfolding 2D image data into 1D sequence and applying RNN units.}
\label{fig:long}
\end{figure}

\subsection{RNN units with Explicit Long-range Conditioning}

\begin{figure}
\begin{center}
   \includegraphics[width=1\linewidth]{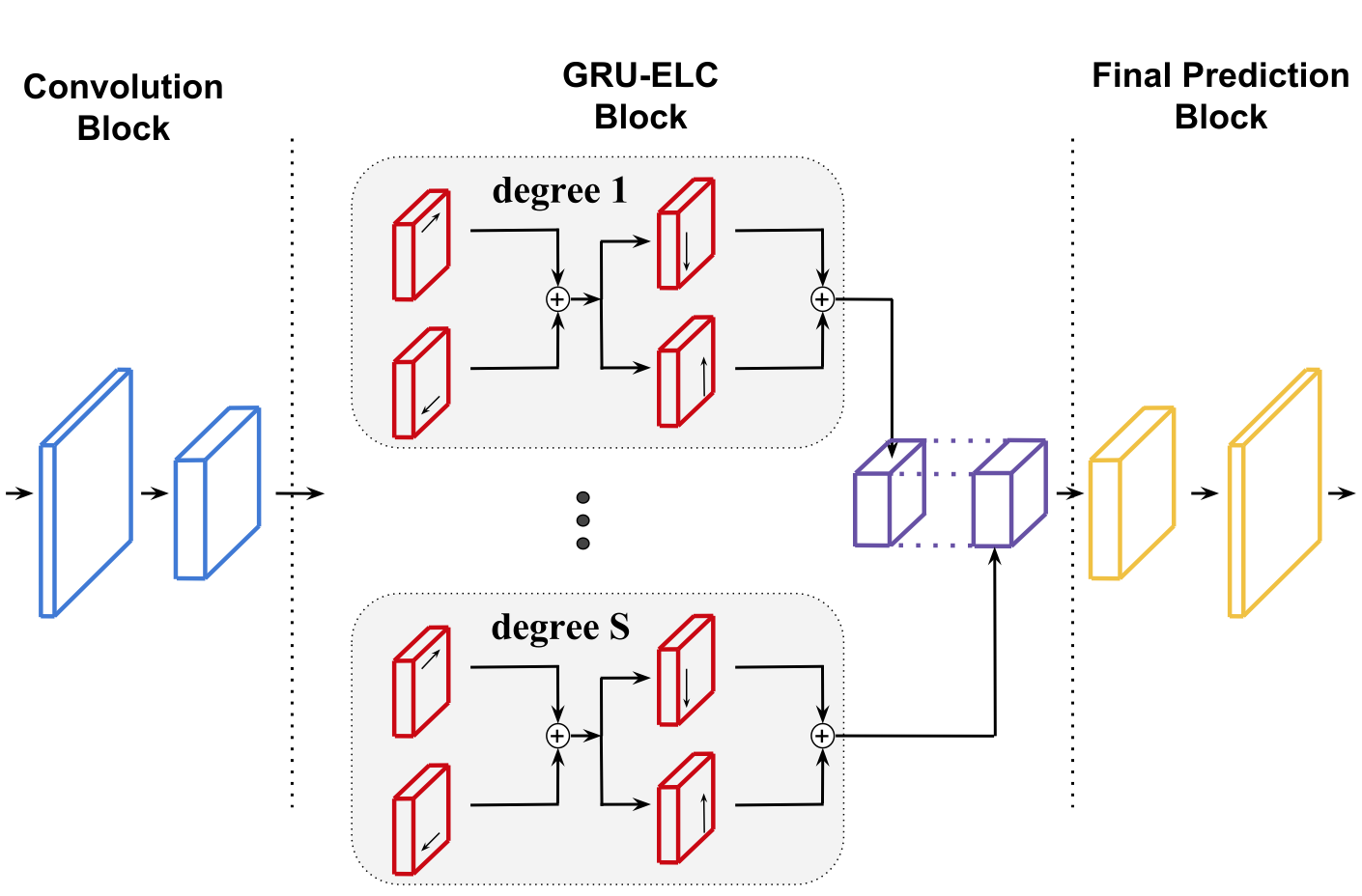}
\end{center}
   \caption{Illustration of our scene labeling algorithm described in Section 4. Blue cuboids denote a convolution layer followed by a pooling layer. Red blocks denote GRU-ELC units with arrow representing the unfolding direction. Purple cuboids denote concatenation operation. Yellow cuboids denote a unpooling layer followed by a convolution layer. In the GRU-ELC block, $S$ parallel branches are used to model contextual dependencies between current variable with its degree $1$, $2$, ..., $s$ neighbors. In each branch, four GRU-ELC units are used to model dependency relations in four different unfolding directions. Note that $\oplus$ denote concatenation operation here.}
\label{fig:frame}
\end{figure}

The consequence of  ``impact vanishing'' problem is that dependencies between spatially related variables are broken. Take the $3\times3$ image grid in Fig.\ref{fig:image_grid} for example. $x^2$ and $x^4$ are both degree 1 neighbors of $x^5$. However, after unfolding it into a 1D sequence in a given direction (from left to right and top to bottom) and applying a RNN to it, $x^5$ is only directly conditioned on variable $x^4$. The information from $x^2$ needs to flow through two more variables and then impacts on $x^5$. From the toy example in previous section, we know that when a large image neighborhood is involved, this impact will vanish in practice.

In order to overcome the ``impact vanishing'' problem and bring back the contextual dependency, we generalize existing RNN units to incorporate with long range conditioning. Mathematically, a vanilla RNN-ELC unit has the following formulation:

\begin{gather}
  h^{t} = \sigma_{h}({x^{t}W_x +  \frac{1}{2} (h^{t-1} + h^{t-s})W_h+b_h} ),
 \label{rnn_elc}
\end{gather}

\noindent
$h^{t-s}$ is the explicit long range conditioning and $s$ is a conditioning skip stride. $\frac{1}{2}$ is a constant term to keep $h^t$ stay in the valid activation area of $\sigma_{h}$. This RNN-ELC unit models the following conditional probability: $P(y^t| x^t, h^{t-1}, h^{t-s}, \theta)$. \textbf{Note that no extra weights are introduced here}.

In the scenario in Fig.\ref{fig:image_grid}, we can set $s=3$ and the resulting graphical model is presented in Fig.\ref{fig:unfold_skip}. Now, the information from $x^2$ flows directly into $x^5$ via a skip connection and the impact from $x^2$ is back now.

In order to test the efficiency of the proposed RNN-ELC units \footnote{It is straightforward to generalize equation (13) for an LSTM/GRU unit. So we don't present them here for space reason}, we use the same data and weights as in previous section and run the toy example again. The conditioning skip stride $s$ is set as 20 here. After repeating each experiment for 20 times, $F^t$ is plotted in the top right plot in Fig.\ref{fig:iv}. Compared with a traditional LSTM/GRU unit, $F^t$ of ELC-LSTM/GRU unit drops faster due to the constant term $\frac{1}{2}$. But, there is a strong peak around time step 20, which does not exist in tradition RNN units. Intuitively, the peak means there is a strong dependency between the $t^{th}$ and $(t-s)^{th}$ variable ($t=20$ in this toy example) in RNN-ELC units now as changing $(t-s)^{th}$ variable has a huge impact on the output of the $t^{th}$ variable.

In real world scenarios, a long range dependency is usually desired. The $t^{th}$ pixel could be treated as related to all $k$ pixels vertically above it. In this case, we need to model the dependency between the $t^{th}$ variable and the $(t-s)^{th}$, $(t-2s)^{th}$, ..., $(t-ks)^{th}$ variables.

The top right plot in Fig.\ref{fig:iv} shows that the conditioning encoded by equation (13) is still not powerful enough as the peak around time step 40 is relatively smaller and $F^t$ decreases to near zero again when $t>60$. So, equation (13) is further generalized as below. $k$ can be called as a conditioning scale.

\begin{gather}
  h^{t} = \sigma_{h}({x^{t}W_x +  H^tW_h+b_h} ) \\
H^t  =   \frac{1}{k+1}  (h^{t-1} + \sum_{i=1}^{k} h^{t-i*s}) 
\end{gather}

Equations (14)-(15) are tested against the toy example again. $s$ is set to be 20 and $k$ is set to be 2 and 3. $F^t$ is reported in the second row in Fig.\ref{fig:iv}. It shows that by explicitly conditioning on related variables in a longer range, the length of valid dependency becomes longer. Next sections show how to exploit this property of RNN-ELC units and build a model for scene labeling which captures a desired long range contextual dependency.

\section{Scene Labeling using Gated Recurrent Units with Explicit Long Range Conditioning}

\begin{figure}[t]
\centering

   \begin{subfigure}{0.45\linewidth}
   \centering
   \includegraphics[width=0.73\linewidth]{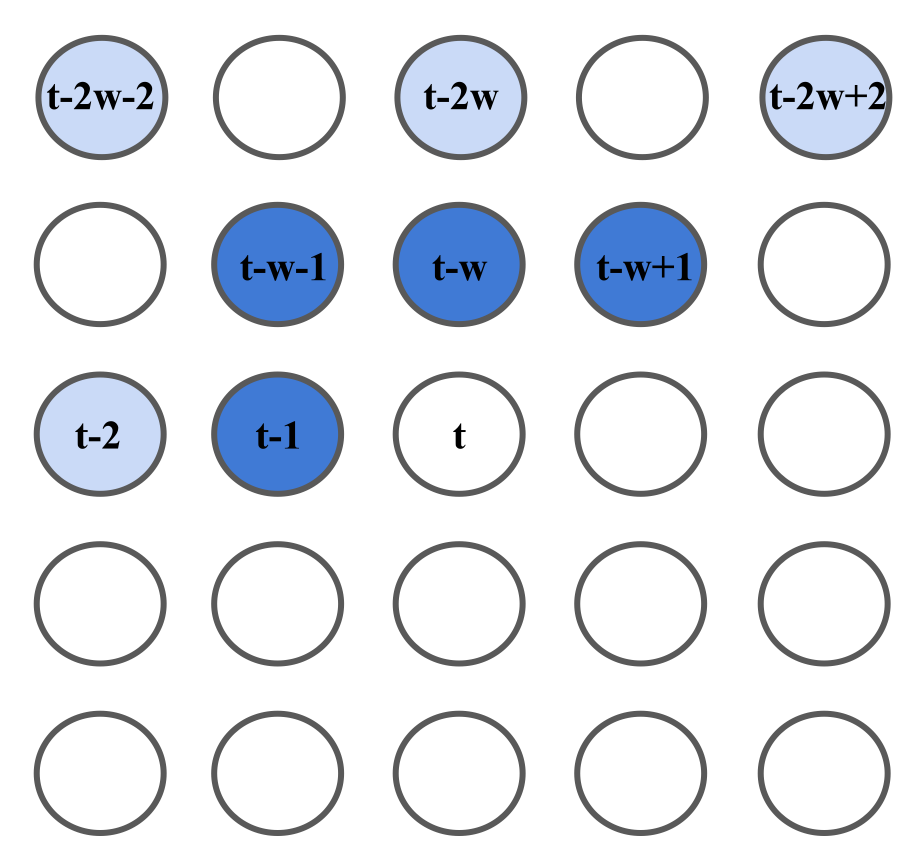}
   \caption{Original 2D grid and dependency relation.}
   \label{fig:image_grid_5} 
\end{subfigure}
\hfill
\begin{subfigure}{0.45\linewidth}
   \centering
   \includegraphics[width=1\linewidth]{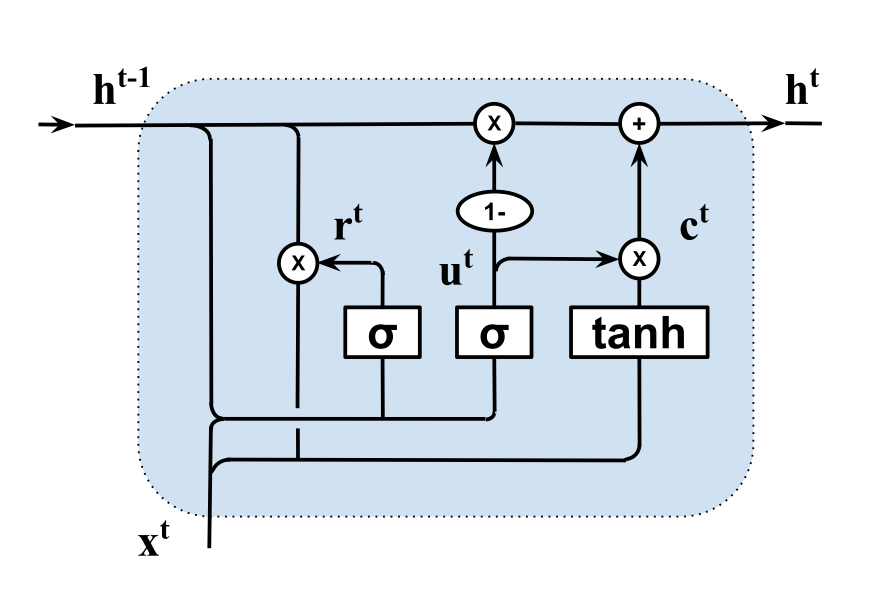}
   \caption{Graphical model of traditional GRU unit.}
   \label{fig:gru}
\end{subfigure}
\begin{subfigure}{1\linewidth}
   \centering
   \includegraphics[width=1\linewidth]{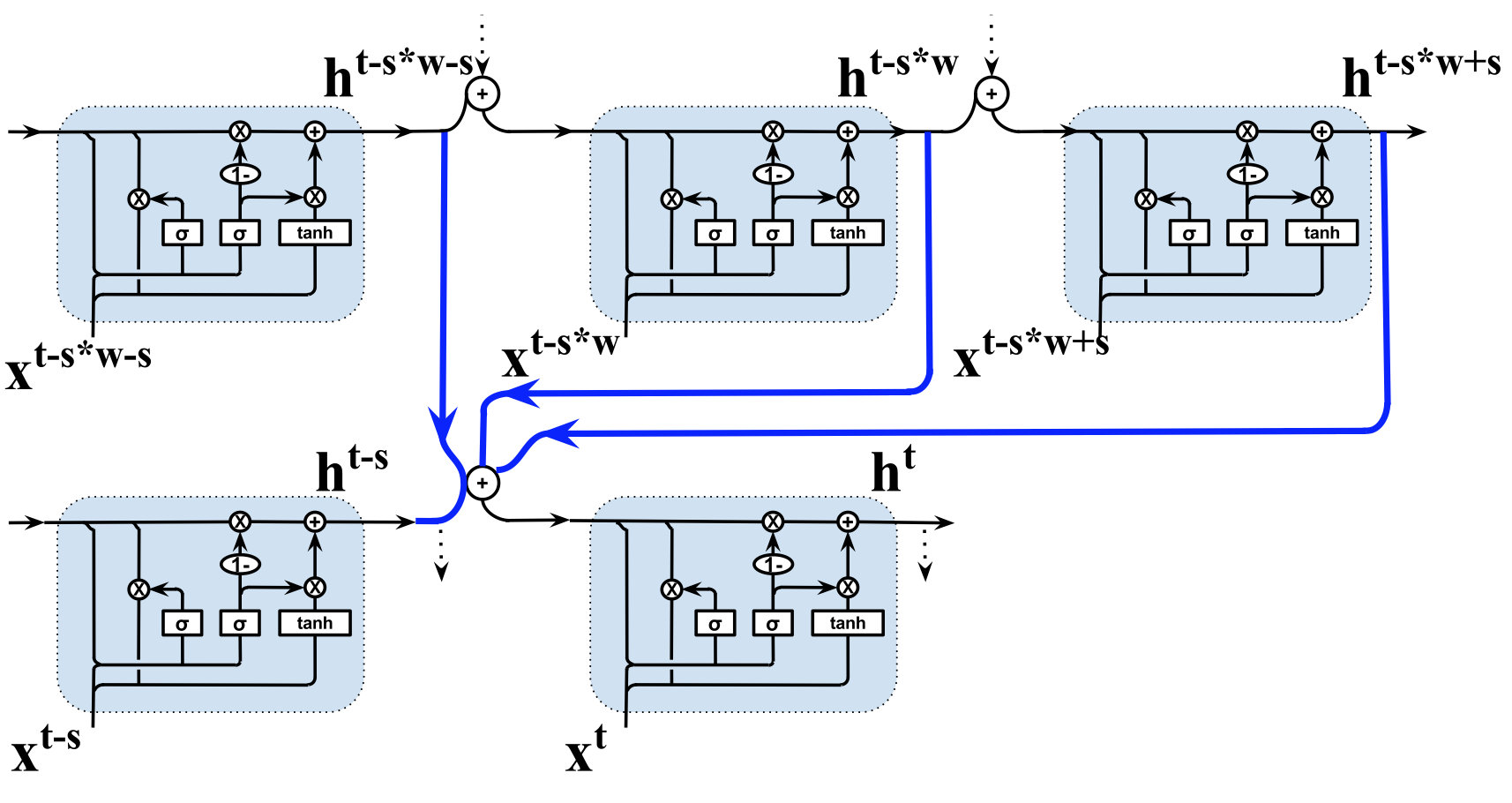}
   \caption{Graphical model of GRU-ELC unit for degree $s$ dependency of the $t^{th}$ variable.}
   \label{fig:gru_elc_s2}
\end{subfigure}

\centering
\caption{Graphical model examples of GRU-ELC units. Note that in (d) only the graphical models of variable $t$ and its degree 2 neighboring variables are drew here. All other variables are ignored as they are not implicitly dependent with variable $t$.  $\sigma$, $tanh$, $\oplus$, \textbf{1-} denote the sigmoid activation, tangent activation, element wise addition, and 1 minus input, respectively. Best viewed with zoom in.}
\end{figure}

An overview of our scene labeling algorithm is presented in Fig.\ref{fig:frame}. It is based on GRU-ELC units\footnote{we choose GRU-ELC units here instead of LSTM-ELC units because GRU-ELC units decay slower than LSTM-ELC units in our toy example as shown in Fig. \ref{fig:iv}} and CNNs. There are three blocks: convolution block, GRU-ELC block, and final prediction block. The convolution block encodes images into features. It is initialized by certain layers from VGG-16 network \cite{vgg}. The final prediction block uses unpooling layers followed by convolution layers to make final predictions of the same resolution as input images. This block is trained from scratch. The GRU-ELC block models contextual dependencies over features.

A multi scale contextual dependency is encoded in our GRU-ELC block. For each variable, its relations with neighbors from degree 1 to degree $S$ are modeled in our framework. Take the 2D grid in Fig. \ref{fig:image_grid_5} for example. Assume the grid is of width $w$ and is unfolded into a 1D sequence from left to right and top to bottom. For variable $t$, its degree 1 neighbors include the $(t-1)^{th}$, $(t-w-1)^{th}$, $(t-w)^{th}$, $(t-w+1)^{th}$ variables and its degree 2 neighbors include the $(t-2)^{th}$, $(t-2w-2)^{th}$, $(t-2w)^{th}$, and $(t-2w+2)^{th}$ variables \footnote{Note that we only consider neighboring variables in diagonal, vertical, and horizontal directions. Because conditioning on too many variables in one RNN-ELC unit will increase $k$ in equation 15 and make the impact decay faster.}.

For dependency in degree $k$, a group of 4 GRU-ELC units is utilized, one GRU-ELC unit for one unfolding direction. Four directions are considered in our framework as inspired by \cite{10, 13, 14}. Take the left-to-right and top-to-bottom direction for example. Following the mechanics as equations (14) - (15), the GRU-ELC unit obeys the following equations.

\begin{gather}
H^t  =   \frac{1}{4}  (h^{t-s} + h^{t- w*s - s} + h^{t - w*s} + h^{t-w*s+s} )  \\
   r^t = \sigma_r (  x^t W_{xr} + H^t W_{hr} + b_r       )    \\
   u^t = \sigma_u (  x^t W_{xu} + H^t W_{hu} + b_u       )    \\   
   c^t = \sigma_c (  x^t W_{xc} +  r^t \odot (  H^t W_{hc} ) + b_c       )    \\
   h^t = (1-u^t) \odot H^t + u^t \odot c^t
\end{gather}

Here $w$ is the width of input grid. For the scenario in Fig. \ref{fig:image_grid_5}, graphical model of GRU-ELC unit used for degree $s$ dependency is presented in Fig. \ref{fig:gru_elc_s2}. The blue lines denote the long range conditioning in GRU-ELC units. They are skip connections which help hidden states from neighboring variables flow directly into current variable. Note that in equation (16) - (20), variable $t$ only conditions on its degree $s$ neighboringing variables. Other neighboring variables will be modeled by other groups of GRU-ELC units in our scene labeling framework. All information will be aggregated together by concatenation operations and then fed into final prediction block.


\newcolumntype{g}{>{\columncolor{Gray}}c}
\begin{table*}
\begin{center}
\label{table:acs}
\begin{tabular}{c|c|c|c|c}
\hline
\multirow{2}{*}{Method} & \multicolumn{2}{c|}{No balancing}  & \multicolumn{2}{c}{ Median balancing} \\

 &   Global (\%) & Class (\%) & Global (\%) & Class (\%) \\



\hline \hline
conv4-decoder & 81.3 & 38.6 & 76.9 & 52.7 \\ 

conv4-GRU & 84.3 & 35.2 & 81.0  & 53.4 \\ 

\hline
\rowcolor{Gray}
conv4-ELC-3 & 86.6 & 45.1 & 83.7 & 61.8 \\ 

\rowcolor{Gray}
conv4-ELC-4 & 87.6 & 47.0 & 84.8 & 62.2 \\ 

\rowcolor{Gray}
conv4-ELC-5 & 87.3 & 46.6 & 85.0 & 62.7 \\

\hline \hline
conv3-decoder & 71.7  & 22.7 & 68.8 & 41.9 \\ 

conv3-GRU & 84.8 & 37.1 & 81.7  &  54.8 \\  

\hline

\rowcolor{Gray}
conv3-ELC-3 & 85.4 & 45.9 & 83.6  &  61.2 \\ 

\rowcolor{Gray}
conv3-ELC-4 & \textbf{87.8} & 46.7 & 84.4 & \textbf{64.0} \\ 

\rowcolor{Gray}
conv3-ELC-5 & 87.4 & \textbf{47.0} & \textbf{85.1}  & 63.6 \\  
\hline

\end{tabular}
\end{center}
\caption{Performance comparison of different choices for GRU-ELC block on SiftFlow. The notation convention is: 1). conv\textbf{X}-decoder denotes this model is built on top of the $convX\_3$ layer of VGG 16 net and no GRU-ELC block is used. Only final prediction block is stacked on top of convolution block; 2). $conv\textbf{X}-GRU$ denotes only traditional GRU units are used in the model; 3) conv\textbf{X}-ELC-\textbf{S} denotes GRU-ELC units are used and the dependencies up to degree $S$ is modeled.}
\label{all_siftflow}
\end{table*}

\section{Experiment}
In order to cover both outdoor and indoor scenes, we select three challenging datasets to test our method, namely the SiftFlow \cite{47}, NYUDv2 \cite{46}, and Stanford Background \cite{48} datasets. Two metrics are used for evaluation, namely global pixel accuracy (\textbf{Global}) and average per-class accuracy (\textbf{Class}). First, a full exploration of different architecture choices is conducted on SiftFlow. Then, state-of-the-art results are presented for NYUDv2, and Stanford Background dataset. 

The convolution block in Fig. \ref{fig:frame} is initialized by certain layers from VGG-16 network \cite{vgg}. Equivalent number of uppooling and convolution layers are used in final prediction block.

All experiments follow a same training protocol. Images with original size are used for training/testing. The training process ends after 25 epochs for all models. The initial learning rate is set as 0.001 and the poly learning policy is adopted to decrease the learning rate after every epoch. All models use recently developed Adam solver \cite{adam} for gradient descent training. The cross entropy loss is used as objective function. Median frequency balancing \cite{2} is applied for comparison. Widely used data augmentation, cropping, flipping, and random jittering are adopted. 

\begin{figure}
\begin{center}
   \includegraphics[width=1\linewidth]{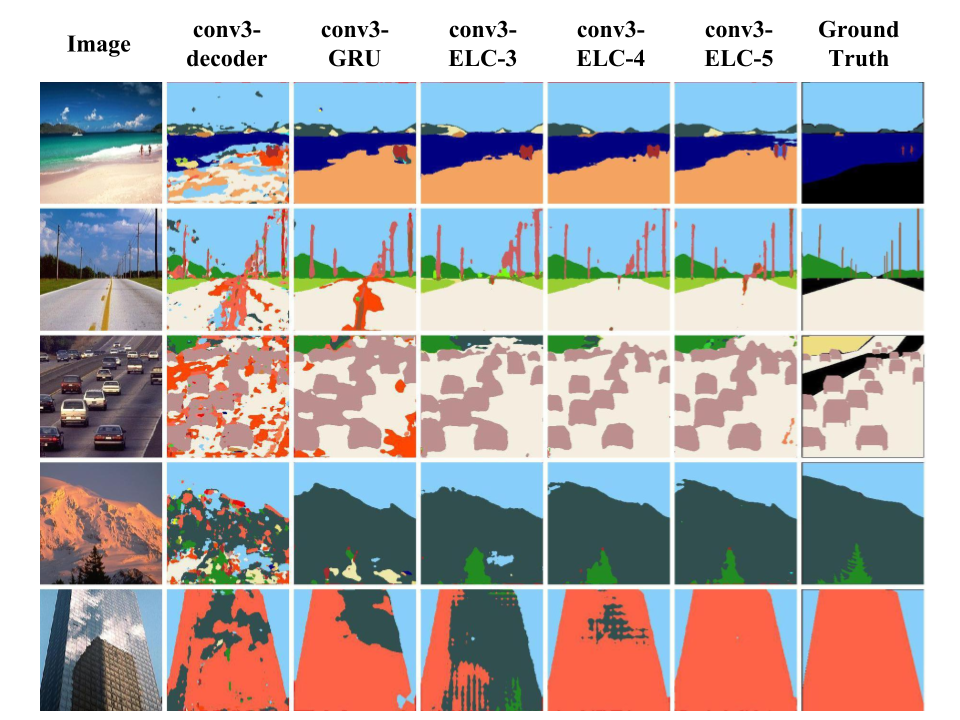}
\end{center}
   \caption{Comparison between models with different settings of GRU-ELC block in our system. All models are trained with median frequency balancing. Note that the black color denotes unknown categories.}
\label{fig:siftflow_conv3_all}
\end{figure}

\subsection{The SiftFlow dataset}
The SiftFlow dataset contains 2,688 images with 33 object categories. All images are of size 256$\times$256 and are captured in outdoor scenes like coast, highway, forest, city, etc. All experiments follow the same 2,488/200 training/testing split as convention.

First, we explore different architecture choices for proposed scene labeling algorithm. We build our system on top of $conv4\_3$ or $conv3\_3$ layer in VGG-16 net. We also tested the $conv5\_3$ layer but the performance is not comparable so we don't report it here. For each feature layer choice, different contextual dependency ranges ($S$ set as 3, 4, and 5) are tested. We also run experiments for architectures without GRU units for comparison. All results are reported in Table \ref{all_siftflow}. 

Experimental results show that GRU-ELC models have superior performances compared with models where only GRU units or no RNN units are used. Some prediction results are visualized in Fig. \ref{fig:siftflow_conv3_all}. Generally, models with longer contextual dependencies have better quantitive and visual performances. This demonstrates that the newly designed GRU-ELC units can effectively model desired long range structure dependencies in images. Specifically, ELC-4 gives much better performance than ELC-3. But ELC-5 only has subtle improvements compared to ELC-4. Considering the fact that ELC-5 requires more computations, ELC-4 actually is a better setting in our system. Note that our models built upon $conv3\_3$ give better results than $conv4\_3$. However, this is not conclusive as we found $conv4\_3$ layer features could give better results on some other datasets. Another observation to notice is that median frequency balancing helps the model achieve better average per-class accuracy but the global pixel accuracy is sacrificed (which has also been found in \cite{5, 2, 1, 36}).

\begin{figure}
\begin{center}
   \includegraphics[width=1\linewidth]{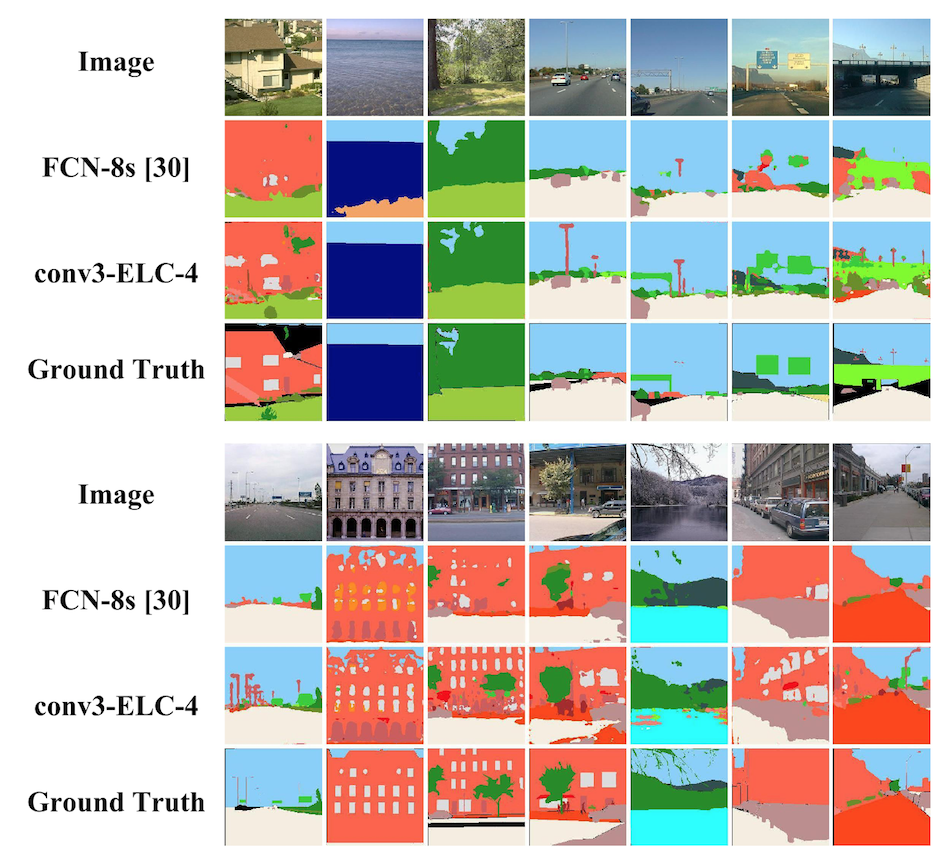}
\end{center}
   \caption{Comparisons of results produced by FCN-8s \cite{3} and conv3-ELC-4 trained with median frequency balancing. conv3-ELC-4 can accurately predict small objects in scenes such the persons, poles, boards, and windows. Note that the black color denotes unknown categories.}
\label{fig:siftflow}
\end{figure}

Our methods are also compared with other state-of-the-art results in Table \ref{siftflow_top}. Comprehensive comparisons show that our methods can outperform both RNN based and other state-of-the-art methods. Especially the average class accuracy has been improved by near 6.3\%. These results show that the proposed algorithm with GRU-ELC units can effectively model long range contextual dependencies in images and thus benefit the scene labeling task a lot. Results produced by our algorithm and FCN-8s \cite{3} are visually compared in Fig.\ref{fig:siftflow}.

\begin{table}
\begin{center}
\label{table:sift}
\begin{tabular}{c|c|c|c}
\hline
\multicolumn{2}{c|}{Method} &   Global (\%) & Class (\%)  \\
\hline\hline
\multicolumn{2}{c|}{Attent to rare class \cite{37}}  &  79.8 &  48.7 \\
\multicolumn{2}{c|}{FCN-16s \cite{3}} & 85.2 & 51.7\\
\multicolumn{2}{c|}{Eigen \textit{et al.} \cite{2}} & 86.8 & 46.4 \\ 
\multicolumn{2}{c|}{Eigen \textit{et al.} \cite{2} \textit{MB}} & 83.8 & 55.7 \\
\multicolumn{2}{c|}{ParseNet \cite{35}} & 86.8 & 52.0 \\
\hline
\multirow{4}{*}{RNN  based} & RCNN \cite{36} & 84.3 & 41.0 \\ 
   & DAG-RNN \cite{16} & 85.3 & 55.7 \\

    & Multi-Path \cite{34} & 86.9 & 56.5 \\

    & Attention \cite{21} & 86.9 & 57.7 \\
\hline
\rowcolor{Gray}
\multicolumn{2}{c|}{conv3-LC-GRU-4} & \textbf{87.8} & 46.7 \\
\rowcolor{Gray}
\multicolumn{2}{c|}{conv3-LC-GRU-4 \textit{MB}}   & 84.4 & \textbf{64.0} \\
\hline
\end{tabular}
\end{center}
\caption{Comparison with state-of-the-art on SiftFlow. Our method is compared against RNN related and other state-of-the-art methods. \textit{MB} denotes this model is trained with median frequency balancing.}
\label{siftflow_top}
\end{table}

\begin{table*}
\begin{center}
\label{table:nyu}
\begin{tabular}{c|c|c|c}
\hline
 Input Information & Method &   Global (\%) & Class (\%)  \\
\hline\hline
RGB & FCN 32s RGB \cite{3} & 60.0 & 42.2\\
\rowcolor{Gray}
RGB & conv4-ELC-4 & \textbf{64.5} & 41.4 \\
\rowcolor{Gray}
RGB & conv4-ELC-4 \textit{MB}   & 62.1 & \textbf{45.5} \\
\hline
RGB+depth & Gupta \textit{et al.}'13 \cite{38}  & 59.1 & 28.4 \\
RGB+depth & Gupta \textit{et al.}'14 \cite{39}  & 60.3 & 35.1 \\
RGB+depth & FCN 32S RGBD \cite{3} & 61.5 & 42.4 \\
RGB+depth & FCN 32S RGB-HHA \cite{3} & 64.3 & 44.9 \\
\rowcolor{Gray}
RGB+depth & FCN 16S RGB-HHA \cite{3} & 65.4 & 46.1 \\
\rowcolor{Gray}
RGB+depth & Wang \textit{et al.} \cite{40} & - & \textbf{47.3} \\
\hline
\rowcolor{Gray}
RGB+depth+normal & Eigen \textit{et al.} \cite{2} & \textbf{65.6} & 45.1 \\
\hline
\end{tabular}
\end{center}
\caption{Comparison with state-of-the-art on NYU. \textit{MB} denotes this model is trained with median frequency balancing.}
\end{table*}

\begin{table}
\begin{center}
\label{table:st}
\begin{tabular}{c|c|c}
\hline
Method &   Global (\%) & Class (\%)  \\
\hline\hline
Sharma \textit{et al.}  \cite{42} & 81.9 & 73.6\\
Mostajabi \textit{et al.}  \cite{41} &  82.1  & 77.3\\
Liang  \textit{et al.} \cite{36} & 83.1 & 74.8 \\
Multi-path \cite{34}  & 86.6 & 79.6 \\
\hline
\rowcolor{Gray}
conv3-ELC-4 & \textbf{87.8} & 81.1 \\
\rowcolor{Gray}
conv3-ELC-4 \textit{MB}   & 84.7 & \textbf{81.5} \\
\hline
\end{tabular}
\end{center}
\caption{Comparison with state-of-the-art on Stanford Background Dataset.\textit{MB} denotes this model is trained with median frequency balancing.}
\end{table}

\subsection{The NYUDv2 dataset}
NYUDv2 is a RGB-D dataset containing 1,449 RGB and depth image pairs for indoor scenes. Standard split contains 795 training images and 654 testing images. 40 categories \cite{38} have been widely used to test the performance of labeling algorithms. Besides RGB images, depth information \cite{3,39} and normal information \cite{2} can also be used for scene labeling. Only RGB images are used in our algorithm in order to get a clear sense about the capacity of our algorithm. The conv4-ELC-4 setting is applied for this dataset. The comparison between our method and other state-of-the-art methods are reported in Table.3. Although only RGB information is used in our method, our results are quite competitive against other methods which additionally use depth or normal information.

\subsection{The Stanford background dataset}
The Stanford background dataset is composed of 715 images with 8 object categories. The images are captured in outdoor scenes and most of them are of size 320$\times$240. Following the standard protocol \cite{36}, a 5-fold cross validation is used for measuring the performance, each of them randomly selects 572 images for training and 143 images for testing. The conv3-ELC-4 setting is adopted for this dataset. Results of our method are reported and compared with other state-of-the-art methods in Table.4. State-of-the-arts results demonstrate the effective of long range dependency for scene labeling problems.


\section{Conclusion}
RNNs are a class of neural network models that have proven effective in modeling internal data dependencies in many areas. There are also many works applying RNNs for image data. In this work, we empirically show that traditional RNN units are not powerful enough to model dependencies in very long sequences due to the ``impact vanishing'' problem. A new RNN unit with Explicit Long-range Conditioning is designed to avoid this problem. Based on the RNN-ELC units, a new scene labeling algorithm is developed in this paper. Various experimental results and comparisons with other state-of-the-art methods demonstrate that our algorithm can effectively capture long range structure dependencies in images and thus give better performances in scene labeling.

Potential directions for future works include: 1). Extend our scene labeling algorithm to take multi-modal input information (like depth or normal information); 2) Apply our new RNN-ELC unit to other image related applications such as image inpainting and image generation.


\appendix
\addcontentsline{toc}{section}{Appendices}
\section*{Appendices}
\section{Network Architecture}

\newcolumntype{g}{>{\columncolor{Gray}}c}
\begin{table*}
\begin{center}
\label{table:acs}
\begin{tabular}{c|c|c|c|c|c|c|c|c|c|c|c|c|c}
\hline \hline

\rowcolor{Gray}
& bird  & sun & bus & pole & boat  \\
\hline
Acc (\%) & 28.3 & 96.5 & 6.0 & 46.8 & 48.8  \\
\hline
Portion (\%) & 0.001 & 0.008 & 0.020 & 0.236 & 0.0333  \\

\hline \hline
\rowcolor{Gray}
& streetlight & sign & staircase & awning & crosswalk\\
\hline
Acc (\%) & 52.8 & 75.1 & 55.4 & 51.9 & 71.0 \\
\hline
Portion (\%)  & 0.046 & 0.062 & 0.080 & 0.081 & 0.083  \\

\hline \hline
\rowcolor{Gray}
& balcony & fence & person & bridge & door  \\
\hline
Acc (\%) & 51.4 & 61.7 & 70.4 & 24.5 & 59.2 \\
\hline
Portion (\%) & 0.102 & 0.150 & 0.163 & 0.193 & 0.367  \\

\hline \hline
\rowcolor{Gray}
& window & sidewalk & rock & sand & car\\
\hline
Acc (\%) & 66.8 & 80.9 & 34.4 & 67.6 & 90.3 \\
\hline
Protion (\%)  & 0.786 & 0.875 & 0.921 & 1.041 & 1.136 \\

\hline \hline
\rowcolor{Gray}
& plant & river & grass & field & sea \\
\hline
Acc (\%) & 48.7 & 84.8 & 75.0 & 53.2 & 84.8 \\
\hline
Portion (\%) & 1.457 & 1.484 & 1.901 & 2.663 & 5.188  \\

\hline \hline
\rowcolor{Gray}
& road & tree & mountain & building & sky\\

\hline
Acc (\%) & 85.0 & 86.6 & 81.4 & 84.0 & 96.7 \\
\hline
Portion (\%)  & 6.952 & 10.846 & 11.762 & 18.400 & 24.434\\

\end{tabular}
\end{center}
\caption{Per-class accuracy achieved by our model ($conv3$-$ELC-4$ $MB$) on the SiftFlow dataset. Categories are organized in ascending order by the data portion in training set.}
\label{per_class_siftflow}
\end{table*}

Here we give details about our scene labeling network. 

The convolution block in our network is initialized by certain layers from VGG-16 network \cite{vgg}. All $conv3$ models are built on top of the $conv3\_3$ layer from VGG-16 network. And all $conv4$ models are built on top of the $conv4\_3$ layer. 

In $conv3$ models, the final prediction block consists of $U-C256-U-C128-C64-Cn$. Here $U$ denotes the up-sampling layer and $CX$ denotes a convolutional layer with $X$ feature maps. $n$ is the number of categories. Kernels of size $3 \times 3$ are used in all convolutional layers except the last one which uses $1\times1$ kernels. Note that 2 up-sampling layers are used here to upscale feature maps to the same resolution of input. In all $conv4$ models, the final prediction block uses the follow architecture: $U-C256-U-C128-U-C128-C64-Cn$.

In the GRU-ELC block, 128 feature maps are used  in each GRU-ELC unit.

\section{Extended Results}
\subsection{Per-class analysis on the SiftFlow dataset}
Here we give a detailed list of per-class accuracy produced by our best model ($conv3$-$ELC$-4 $MB$ ) on SiftFlow dataset. They are organized in Table \ref{per_class_siftflow} in ascending order by the data portion in training set. For example, there are 24.434\% of the training data (pixels) are \textit{sky} but only 0.001\% are \textit{bird}.

Generally, the model performs quite well on categories with large amount of training data like \textit{sky}, \textit{building}, \textit{mountain}, \textit{tree}, \textit{road}, etc. Moreover, our model can also give reasonable performances on some categories where only limited training data are available. For example, our model has a 96.5\% accuracy on $sun$ even when there are only 0.008\% of training data are $sun$. On some other rare categories, such as \textit{sign}, \textit{crosswalk}, \textit{person}, \textit{window}, \textit{sidewalk}, and \textit{sand}, we have accuracies above average. We argue that there are two main reasons for the good per-class performances. Firstly, our model can effectively capture long range contextual dependencies in images and thus yields a good performance on small objects which are usually rare categories  (such as the persons and poles shown in Fig. 6 in the main text). Secondly, the median frequency balancing is used in our $conv3$-$ELC$-4 $MB$ model, which helps to improve performances in rare categories, as also shown in \cite{5, 2, 1, 36}.

\subsection{Visual results on the NYUDv2 and Stanford Background dataset}
\begin{figure*}
\begin{center}
   \includegraphics[width=0.8\linewidth]{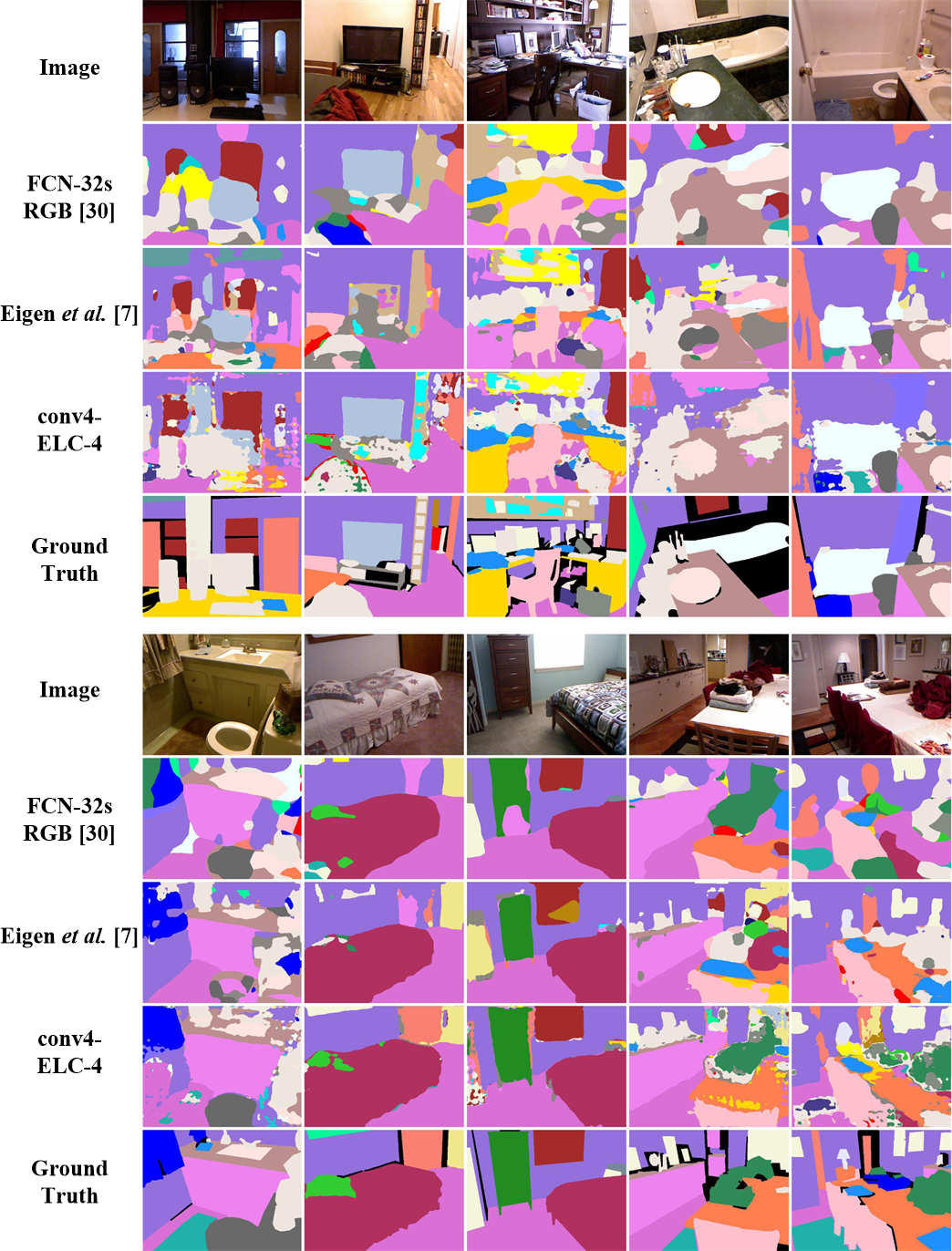}
\end{center}
   \caption{ Comparisons of results produced by FCN-8s \cite{3}, Eigen \textit{et al.} \cite{2}, and conv4-ELC-4 trained with median frequency balancing. Note that the black color denotes unknown categories. }
\label{fig:nyu}
\end{figure*}

Some predictions on the NYUDv2 dataset produced by our $conv4$-$ELC$-4 $MB$ model are visualized in Fig. \ref{fig:nyu}. Results produced by \cite{2} and \cite{3} are compared in Fig. \ref{fig:nyu} as well. From Fig. \ref{fig:nyu} we can see that our model performs much better in local details than other two models in these samples, such as the monitors in the first sample image, the TV screen in the second sample image, and the objects on the desk in the third sample image.

\begin{figure*}
\begin{center}
   \includegraphics[width=1\linewidth]{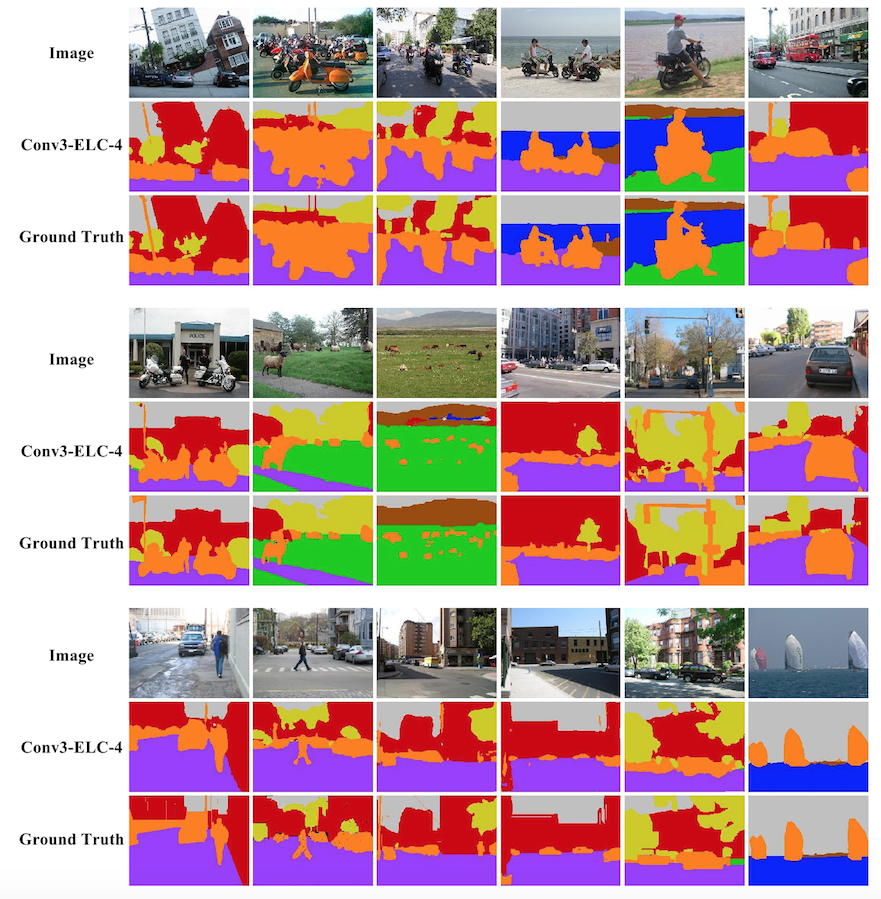}
\end{center}
   \caption{ Visualization of prediction results on the Stanford background dataset produced by our $conv3$-$ELC$-$4$ $MB$ model. }
\label{fig:stanford}
\end{figure*}

Fig. \ref{fig:stanford} presents some samples of our results on the Stanford background dataset. These results show that our model can effectively detect boundaries and accurately segment small objects in images, such as the poles, animals, persons, and vehicles in images.

{\small
\bibliographystyle{ieee}
\bibliography{egbib}
}

\end{document}